\documentclass[10pt]{article} % For LaTeX2e

\usepackage[accepted]{tmlr}
\usepackage{amsmath,amsfonts,bm}

\def\eqref#1{equation~\ref{#1}}
\def\Eqref#1{Equation~\ref{#1}}
\def\1{\bm{1}}

\DeclareMathAlphabet{\mathsfit}{\encodingdefault}{\sfdefault}{m}{sl}
\SetMathAlphabet{\mathsfit}{bold}{\encodingdefault}{\sfdefault}{bx}{n}

\def\sN{{\mathbb{N}}}

\newcommand{\R}{\mathbb{R}}

\DeclareMathOperator*{\argmax}{arg\,max}

\DeclareMathOperator{\sign}{sign}

\usepackage{hyperref}
\usepackage{url}
\title{Cross-client Label Propagation for Transductive and \newline Semi-Supervised Federated Learning}

\author{\name Jonathan Scott \email jonathan.scott@ist.ac.at \\
      \addr ISTA (Institute of Science and Technology Austria), Klosterneuburg, Austria
      \AND
      \name Michelle Yeo \email michelle.yeo@ist.ac.at \\
      \addr ISTA (Institute of Science and Technology Austria), Klosterneuburg, Austria
      \AND
      \name Christoph H. Lampert \email chl@ist.ac.at\\
      \addr ISTA (Institute of Science and Technology Austria), Klosterneuburg, Austria}

\def\month{10}  % Insert correct month for camera-ready version
\def\year{2023} % Insert correct year for camera-ready version

\def\openreview{\url{https://openreview.net/forum?id=gY04GX8R5k}} % Insert correct link to OpenReview for camera-ready version

\usepackage{microtype}
\usepackage{graphicx}
\usepackage{booktabs} % for professional tables
\usepackage{enumitem}

\usepackage{amsthm, amsmath, amssymb, amsfonts, mathtools}
\usepackage{bbm}
\usepackage{xspace}

\usepackage[font=normalsize]{subfig}
\usepackage{caption}

\usepackage[ruled, linesnumbered, vlined]{algorithm2e}
\usepackage{algcompatible}

\usepackage{cleveref}

\newcommand{\trans}{\top}

\algrenewcommand{\algorithmicrequire}{\textbf{Input:}}
\algrenewcommand{\algorithmicensure}{\textbf{Output:}}
\algrenewcommand\algorithmiccomment[1]{\hfill//~\textit{#1}}
\SetKwInput{Input}{Input}%
\SetKwInput{Output}{Output}%
\SetCommentSty{textit}%
\SetAlgoNlRelativeSize{0}
\DontPrintSemicolon

\makeatletter
\DeclareRobustCommand\onedot{\futurelet\@let@token\@onedot}
\def\@onedot{\ifx\@let@token.\else.\null\fi\xspace}
\def\iid{{i.i.d}\onedot}
\def\eg{{e.g}\onedot} 
\def\ie{{i.e}\onedot}

\makeatother

\theoremstyle{definition}

\newcommand{\methodname}{XCLP}
\newcommand{\Fedprop}{FedAvg+\method}

\newcommand{\method}{\methodname\xspace}

\newcommand{\oursum}{SecureRowSums\xspace}

\newcommand{\myparagraph}[1]{\noindent\textbf{#1}\xspace}

\begin{document}
\maketitle

\begin{abstract}
We present \emph{Cross-Client Label Propagation (\method)}, a new method 
for transductive and semi-supervised federated learning. 
\method estimates a data graph jointly from the data of multiple clients 
and computes labels for the unlabeled data by propagating label information 
across the graph. 
%n
To avoid clients having to share their data with anyone, \method 
employs two cryptographically secure protocols: 
\emph{secure Hamming distance computation} and \emph{secure summation}.
We demonstrate two distinct applications of \method within federated learning. 
In the first, we use it in a one-shot way to predict labels for 
unseen test points. In the second, we use it to repeatedly 
pseudo-label unlabeled training data in a federated semi-supervised 
setting.  
Experiments on both real federated and standard benchmark datasets 
show that in both applications \method achieves higher classification 
accuracy than alternative approaches.
\end{abstract}

% SECTION 1

\section{Introduction}
Federated Learning (FL)~\citep{fedavg} is a machine learning paradigm in 
which multiple clients, each owning their own data, cooperate to jointly solve a learning task. 
The process is typically coordinated by a central server.
The defining restriction of FL is that client data must remain on device 
and cannot be shared with either the server or other clients.
In practice this is usually not due to the server being viewed as a hostile 
party but rather to comply with external privacy and legal constraints that 
require client data to remain stored on-device.
To date, the vast majority of research within FL has been focused 
on the supervised setting, in which client data is fully 
labeled and the goal is to train a predictive model. 
In this setting a well-defined template has emerged: first proposed 
as \emph{federated averaging}~\citep{fedavg}, this consists of 
alternating between local model training at the clients and model aggregation at the server.

However, in many real-world settings fully labeled data may not be available. For instance, in \emph{cross-device} FL, smartphone users are not likely to be interested in annotating more than a handful of the photos on their devices \citep{flair}. %
Similarly, in a \emph{cross-silo} setting the labeling of medical imaging data may be both costly and time consuming~\citep{owkin_selfsup}. 
As such, in recent years there has been growing interest in developing algorithms that can learn from partly labeled or fully unlabeled data in a federated setting~\citep{advances_FL}.
For such algorithms it can be beneficial, or even essential, to go beyond the standard federated framework of 
model-centric learning and develop techniques that directly leverage client data interactions. 
Examples include federated clustering~\citep{fedkmeans} and dimensionality reduction~\citep{fedpca}, 
where clients compute statistics based on their data and the server computes with aggregates of these statistics.

\myparagraph{Contribution}\quad 
In this work we propose \emph{Cross-Client Label Propagation (\method)} which follows a data-centric approach. 
\method allows multiple clients, each with labeled and unlabeled data, to cooperatively compute labels for their 
unlabeled data. This is done using a transductive approach, where the goal is label inference restricted to some 
predefined set of unlabeled examples. In particular this approach does not require a model to be trained in order 
to infer labels.
Specifically, \method takes a graph-based approach to directly assign labels to the unlabeled data.
It builds a joint data graph of a group of clients and propagates the clients' label information along the edges. 
Naively, this approach would require the clients to centrally share their data.
That, however, would violate the constraints of federated learning.

\method allows for multiple clients to jointly infer labels from a cross-client data graph \emph{without} them having to share their data. We refer to this privacy guarantee, in which a client's data never leaves their 
own device, as data confidentiality.
To achieve this \method exploits the modular and distributed nature of the problem. % 
It uses locality-sensitive hashing and secure Hamming distance computation to efficiently estimate the cross-client data graph.
It then distributes the label propagation computation across clients and aggregates the results using a customized variant of \textit{secure summation}. 
The key benefits of this approach are:
\begin{itemize} %[topsep=0pt,itemsep=2pt,partopsep=4pt, parsep=4pt]
    \item \method enables the data of multiple clients to be leveraged when estimating the graph and propagating labels. 
    This is beneficial as the prediction quality of label propagation increases substantially when more data is used.
    \item \method achieves data confidentiality, since it does not require clients to share their data with anyone else. Instead clients only have to share approximate data point similarities with the server.
    \item \method is communication efficient as it does not require training a model over multiple communication 
    rounds but requires only a single round of communication.
\end{itemize}
As a technique to transfer label information from labeled to unlabeled points, \method is versatile enough to be used in a variety of contexts. 
We illustrate this by providing two applications within federated learning. 
In the first we employ \method purely for making predictions at inference time. We demonstrate empirically on a real-world, highly heterogeneous, federated medical image dataset that \method is able to assign high quality labels to unlabeled data. When all clients are partly labeled we observe \method to outperform purely local label propagation, which illustrates the benefits of leveraging more data. \method also obtains strong results when using fully labeled clients to infer labels on different, fully unlabeled clients, even when these clients have very different data distributions. In both these scenarios we find that running \method on features obtained from models trained using \emph{FederatedAveraging} gives significantly better accuracy than purely using the model predictions.

In the second application we tackle the problem of federated semi-supervised learning. In this scenario clients possess only partly labeled data and the goal is to train a classifier by leveraging both labeled and unlabeled client data. In this setting we employ \method in the training process by integrating it into a standard federated learning pipeline. Specifically, during each round of \emph{FederatedAveraging} we use \method to assign pseudo-labels and weights to the unlabeled data of the current batch of clients. These are then used to train with a weighted supervised loss function. 
Our experiments show that this pseudo-labelling approach outperforms all existing methods for federated semi-supervised learning, as well as a range of natural baselines in the standard federated CIFAR-10 benchmark. 
Going beyond prior work, we also evaluate on more challenging datasets, namely CIFAR-100 and Mini-ImageNet, where we also observe substantial improvements in accuracy.

% SECTION 2

\section{Related Work}
\myparagraph{Federated Learning}\quad % => SSL
Federated learning (FL) \citep{fedavg} was originally proposed for learning on 
private fully labeled data split across multiple clients. 
For a survey on developments in the field see \citep{advances_FL}.
A number of recent works propose federated learning in the absence of fully labeled data.
Methods for cluster analysis and dimensionality 
reduction have been proposed~\citep{fedkmeans,fedpca}, in which the server
acts on aggregates of the client data, as opposed to client models.
Other works have focused on a model based approach to federated semi-supervised learning (SSL).
\citet{FedMatch} propose inter-client consistency and parameter decomposition to separately learn from labeled and unlabeled data.
\citet{fedsiam} apply consistency locally through client based teacher models.
\citet{RSCFed} combine local semi-supervised training with an enhanced aggregation scheme which re-weights client models based on their distance from the mean model.
\citet{grad_diversity} and \citet{semiFL} focus on a setting in which the 
server has access to labeled data. In this setting \citet{grad_diversity} combine local consistency with grouping of client 
updates to reduce gradient diversity while \citet{semiFL} combine consistency, through strong data augmentation, with pseudo-labeling unlabeled client data.
Our approach to federated SSL is to iteratively apply \method to pseudo-label unlabeled client data. This approach differs from prior work by making use of data interactions between multiple clients to propagate label information over a cross client data graph.  
Related to this idea is the notion of federated learning on graphs~\citep{graph1,graph2,graph3}. However, these works are primarily interested in learning from data that is already a graph. In contrast \method estimates a graph based on similarities between data points, in order to spread label information over the edges.

\myparagraph{Label Propagation}\quad
Label Propagation~\citep{LP_older,label_prop} was originally proposed as a tool for transductive learning
with partly labeled data. Over time it has proven to be a versatile tool for a wide range of problems.
Several works have applied LP to the problem of domain adaptation. \citet{few_shot_LP} apply LP over a learned graph 
for few shot learning.
\citet{LP_subpop_shift} develop a framework for domain adaptation by combining LP with a teacher trained on the source.
\citet{khamis2014coconut} use LP as a prediction-time regularizer for collective classification tasks.
In the context of deep semi-supervised learning \citet{deep_label_prop} make use of LP as a means
of obtaining pseudo-labels for unlabeled data which are then used in supervised training.
Several works apply LP to problems with graphical data. \citet{LP_and_simple_models} observe that 
combining linear or shallow models with LP can lead to performance that is on par with or better than
complex and computationally expensive GNNs.
\citet{GNN_LP} apply LP as a regularizer for graph convolutional neural networks when learning edge weights and quantify 
the theoretical connection between them in terms of smoothing.

%\SetCustomAlgoRuledWidth{\linewidth}
\newcommand{\XSstep}{\SetNlSty{}{}{-XS:}}
\newcommand{\CSstep}{\SetNlSty{}{}{-CS:}}
\newcommand{\SSstep}{\SetNlSty{}{}{-SS:}}
\IncMargin{3.5em}
\begin{algorithm*}[t]
\Crefname{algorithm}{Alg.}{Algs.}
\caption{\texttt{Cross-ClientLabelPropagation}}
\label{alg:crossclientLP}
\Indm\Indm % \alpha?
\Input{set of participating clients $P$, client data $(V^{(j)}, Y^{(j)})_{j\in P}$ \tcp*{data stored on-device at clients}}
%\Input{server, clients $P$ with data $(V^{(j)},  Y^{(j)})_{j\in P}$}
\Indp\Indp
%\color{red}
\XSstep 
Setup: clients exchange private and public keys, agree on random seed $s$\label{xcLPline:seed}\;
%\color{blue}
\CSstep
\For{client $j \in P$ in parallel}{
$\Pi\in\R^{L\times d} \quad\text{with}\quad \Pi_{ij}\stackrel{\iid}{\sim} \mathcal{N}_{\text{seed}=s}(0,1)$
\label{xcLPline:Pi} \hfill  \tcp*{same $\Pi$ for each client}
$B^{(j)} \gets \sign(\Pi\,(V^{(j)})^\top)$\label{xcLPline:Bj}  \tcp*{LSH projection}
} % end of For
%\color{red}
\XSstep
$H \gets \texttt{SecureHamming}((B^{(j)})_{j\in P})$ \label{xcLPline:secure_hamming} \tcp*{server gets Hamming matrix}
%\color{olive}
\SSstep
$A \gets \cos(\frac{\pi}{L}H)$ \tcp*{estimate cosine similarity matrix \label{xcLPline:A}}
$B \gets \text{sparsify}(A)$ \tcp*{keep $k$ largest entries per row, set others to $0$\label{xcLPline:B}}
$W = B + B^\trans$ \tcp*{symmetrize\label{xcLPline:W}}
$\mathcal{W} \gets D^{-\frac{1}{2}} W D^{-\frac{1}{2}} \text{ for } D=\operatorname{diag}(d_1,\dots,d_n) \text{ with } d_i=\sum_{j}W_{ij}$
\tcp*{normalize\label{xcLPline:curlyW}}
$S \gets (\text{Id}_{n\times n} - \alpha \mathcal{W})^{-1} $   \label{xcLPline:S}
\tcp*{influence matrix}
%\color{blue}
\CSstep
\For{client $j \in P$ in parallel}{
$S_L^{(j)} \gets \text{labeled-cols}_j(S)$ \tcp*{client gets columns corresponding to labeled data}   \label{xcLPline:S-j}
$\bar Z^{(j)} \gets S_L^{(j)} Y_L^{(j)}$ \tcp*{compute local contribution to overall label propagation\label{xcLPline:Zj}}
\XSstep
$Z^{(j)} \gets \texttt{SecureRowSums}\big((\bar Z^{(k)})_{k\in P}\big)_j$ \tcp*{client gets its part of aggregated contributions \label{xcLPline:secure_sum}}
\CSstep
%$\hat{Z}^{(j)} \leftarrow \text{normalize}(Z^{(j)})$    \label{xcLPline:normalize}
%\tcp*{line \ref{LPline:normalizeZ} of \Cref{alg:LP}} 
$\hat y^{(j)} \gets \big(\argmax_{c=1,\dots,C} Z^{(j)}_{i,c}\big)_{i=1,\dots,n^{(j)}}$ \tcp*{predict labels\label{xcLPline:pseudo}}
$\omega^{(j)} \gets \Big(1-\texttt{entropy}\big(\frac{(Z^{(j)}_{i,c})_{c=1,\dots,C}}{\sum_{c=1}^C Z^{(j)}_{i,c}}\big)/\log C\Big)_{i=1,\dots,n^{(j)}}$ \tcp*{predicted label confidences\label{xcLPline:omega}}
}
\Indm\Indm
\Output{predicted labels and confidences $(\hat{y}^{(j)},\omega^{(j)})_{j\in P}$ 
\tcp*{available only to respective clients}}
\Indp\Indp
\end{algorithm*}

% SECTION 3
\section{Method}\label{sec:method}
In this section we begin by introducing the problem setting. In \Cref{subsec:method} we present our method \method, in \Cref{subsec:crypto}
we describe the cryptographic protocols we make use of and in \Cref{subsec:analysis} we provide an analysis of \method.

Let $P$ be a set of client devices for which labels should be propagated. 
Note that we place no restrictions on $P$. It could be all clients in a federated learning scenario,
a randomly chosen subset, or a strategically chosen subset, \eg based on client similarity, diversity 
or data set sizes.

Each client $j\in P$ possesses a set of $n^{(j)}$ $d$-dimensional data vectors, of which $l^{(j)}$ are labeled, 
\ie $V^{(j)} =  (v^{(j)}_1, \dots v^{(j)}_{l^{(j)}}, v^{(j)}_{l^{(j)} + 1}, \dots v^{(j)}_{n^{(j)}})^\top\in\R^{n^{(j)}\times d}$,
with partial labels $\{y^{(j)}_1,\dots,y^{(j)}_{l^{(j)}}\}$ from $C$ classes, which we encode in zero-or-one-hot matrix form: 
$Y^{(j)}\in\{0,1\}^{n^{(j)}\times C}$ with $Y^{(j)}_{ic} = \mathbbm{1}\{y^{(j)}_i \!\!= c\}$ for $1\leq i\leq l^{(j)}$ and $Y^{(j)}_{ic}=0$ otherwise.
Note that this setup includes the possibility for a client to have only labeled data, $n^{(j)} = l^{(j)}$ or only unlabeled data, $l^{(j)} = 0$. 
We denote the total amount of data by $n\coloneqq \sum_{j\in P}n^{(j)}$.
Our goal is to assign labels to the unlabeled data points, \ie, \emph{transductive learning}~\citep{vapnik1982}.
The process is coordinated by a central server, which we 
assume to be \emph{non-hostile}. 
That means, we trust the server to operate on non-revealing 
aggregate data and to return correct results of computations.\footnote{In particular we exclude \emph{malicious} servers in the cryptographic sense that would, \eg, be 
allowed to employ attacks such as model poisoning or generating 
fake clients in order to break the protocol.}
At the same time, we treat clients and server as \emph{curious},
\ie, we want to prevent that at any point in the process 
any client's data is revealed to the server, or to any other client.

Our main contribution in this work is, \emph{Cross-Client Label Propagation (\method)}, 
an algorithm for assigning labels to the unlabeled data points. \method works
by propagating label information across a neighborhood graph that is built jointly 
from the data of all participating clients without revealing their data.
Before explaining the individual steps in detail,
we provide a high level overview of the method.

\method consists of three phases:
1) the clients jointly compute similarity values between 
all of their data points and transfer them to the server,
2) the server uses these similarities to construct a neighborhood graph and 
infers an influence matrix from this, which it distributes 
back to the clients,
3) the clients locally compute how their data influences 
others, aggregate this information, and infer labels for 
their data.

The key challenge is how to do these steps without
the clients having to share their data and labels with each 
other or with the server.
\method manages this by formulating the problem in a 
way that allows us to use only light-weight cryptographic 
protocols for the steps of computing similarities and 
aggregating label information.

\subsection{Cross-Client Label Propagation (\method)}\label{subsec:method}
Algorithm~\ref{alg:crossclientLP} shows pseudocode for \method.
To reflect the distributed nature of \method we mark the execution type
of each step:
\emph{client steps (CS)} are steps that clients 
do locally using only their own data, 
\emph{server steps (SS}) are steps that the 
server executes on aggregated data, 
\emph{cross steps (XS)} are steps that require
cross-client or client-server interactions.

As a setup step (line \ref{xcLPline:seed}) the 
clients use a secure key exchange procedure to 
agree on a shared random seed that remains unknown 
to the server. 
This is a common step in FL when cryptographic 
methods, such as \emph{secure model aggregation}, are 
employed, see \citet{BonawitzIKMMPRS17}.

\myparagraph{Phase 1.} 
The clients use the agreed-on random seed to generate a common 
matrix $\Pi\in\R^{L\times d}$ with unit Gaussian random entries (line \ref{xcLPline:Pi}).
Each client, $j$, then multiplies its data matrix $V^{(j)}$ by 
$\Pi$ and takes the component-wise sign of the result, thereby 
obtaining a matrix of $n^{(j)}$ $L$-dimensional binary vectors, 
$B^{(j)}$ (line \ref{xcLPline:Bj}).  %  matrix 
In combination, both steps constitute a local 
\emph{locality-sensitive hashing (LSH)}~\citep{indyk1998approximate}
step for each client.
A crucial property of this encoding is that the (cosine) similarity 
between any two data vectors, $v,v'$, can be recovered from their binary 
encodings 
$b,b'$:
$\operatorname{sim}(v,v') := \frac{\langle v, v'\rangle}{\|v\|\|v'\|} \approx \cos( \pi h(b, b')/L),$
where 
$h(b, b')=\sum_{l=1}^L b_l \oplus b'_l$ 
is the Hamming distance (number of bits that differ) between binary vectors and 
$\oplus$ is the XOR-operation.
Since all clients use identical random projections, 
this identity holds even for data points located on different clients. 
See \Cref{app:LSH} for details on LSH.

In line~\ref{xcLPline:secure_hamming} the necessary Hamming distance 
computations take place using a cryptographic subroutine that we detail 
in Section~\ref{subsec:crypto}.
Note that cryptographic protocols operate most efficiently on 
binary vectors, and Hamming distance is particularly simple 
to compute. In fact, this is the reason why we transform 
the data using LSH in the first place. 
Ultimately, from this step the server obtains the matrix of all 
pairwise Hamming distances, $H\in\mathbb{Z}^{n\times n}$, but no
other information about the data.

\myparagraph{Phase 2.} Having obtained $H$ the server executes a 
number of steps by itself. 
First, it converts $H$ to a (cosine) similarity matrix 
$A\in\R^{n\times n}$ (line \ref{xcLPline:A}). 
It sparsifies each row of $A$ by keeping the $k$ 
largest values and setting the others to $0$ (line \ref{xcLPline:B}).
From the resulting matrix, $B$, it constructs a weighted 
adjacency matrix of the data graph, $\mathcal{W}$ by 
symmetrization (line~\ref{xcLPline:W})
and normalization (line~\ref{xcLPline:curlyW}).

If not for the aspect of data confidentiality, we could now 
achieve our goal of propagating label information 
along the graph edges from labeled to unlabeled points
in the following way: form the concatenation of 
all partial label matrices, $Y=(Y^{(j)})_{j\in P}\in\{0,1\}^{n\times C}$, 
and compute $Z=SY\in\R^{n \times C}$, where $S=(\operatorname{Id}-\alpha\mathcal{W})^{-1}$ 
is the \emph{influence matrix}, and $\alpha\in(0,1)$ is a hyperparameter.
See~\Cref{app:LP} for an explanation how this 
step corresponds to the propagation of labels over the graph.  

\method is able to perform the computation of the 
\emph{unnormalized class scores}, $Z$, without having 
to form $Y$, thereby preserving the confidentiality of the labels. 
Instead, it computes only the influence matrix, $S$, centrally 
on the server (line~\ref{xcLPline:S}), while the multiplication 
with the labels will be performed in a distributed way across 
the clients.

\myparagraph{Phase 3.}
Observe that the computation of $Z$ can also be written as 
$Z = \sum_{j \in P} S^{(j)} Y^{(j)}$,
where $S^{(j)} \in \R^{n\times n^{(j)}}$ is the sub-matrix 
of $S$ consisting of only the columns that correspond to 
the data of client $j$.
We can refine this further, note that all rows of $Y^{(j)}$ 
that correspond to the unlabeled data of client $j$ are 
identically $0$ by construction and hence do not 
contribute to the multiplication. 
Therefore, writing $Y_L^{(j)} \in \R^{l^{(j)} \times C}$ 
for the rows of $Y^{(j)}$ that correspond to labeled points 
and $S_L^{(j)}\in \R^{n\times l^{(j)}}$ for the corresponding 
columns of $S^{(j)}$, it also holds that $Z = \sum_{j \in P} S_L^{(j)} Y_L^{(j)}$.

Using this observation, Algorithm~\ref{alg:crossclientLP} 
continues by each client $j$ receiving $S_L^{(j)}$ from the server 
(line \ref{xcLPline:S-j}). It then locally computes 
$\bar Z^{(j)} = S_L^{(j)} Y_L^{(j)} \in \R^{n \times C}$ (line \ref{xcLPline:Zj}), 
which reflects the influence of $j$'s labels on all other data points.

By now, the clients have essentially computed $Z$, but the result 
is additively split between them: $Z = \sum_{j\in P}\bar Z^{(j)}$. 
To compute the sum while preserving data confidentiality, \method uses a secure 
summation routine (line~\ref{xcLPline:secure_sum}), as commonly used in FL for model averaging~\citep{BonawitzIKMMPRS17}.
However, to increase its efficiency we tailor it to the task, 
see \Cref{subsec:crypto} for details. 
As a result, each client $j$ receives only those rows of $Z$ that 
correspond to its own data, $Z^{(j)} \in \R^{n^{(j)}\times C}$ (line~\ref{xcLPline:Zj}).
From these, it computes labels and confidence values for its data (lines \ref{xcLPline:pseudo}-\ref{xcLPline:omega}).

\subsection{Cryptographic Subroutines}\label{subsec:crypto}

In this section we describe the proposed cryptographic 
protocols for \emph{secure summation} and \emph{secure Hamming 
distance computation} that allow clients to run \method without
having to share their data. 

% floating point: aliasgariBZS13
% fixed point: catrina2010secure
\texttt{\bfseries\oursum}\quad
We propose a variation of the \emph{secure summation} that is commonly used in FL~\citep{BonawitzIKMMPRS17}.
For simplicity we describe the case where all values belong to $\mathbb{Z}_l$ 
for some $l \in \mathbb{N}$, though extensions to fixed-point or floating-point 
arithmetic also exist~\citep{catrina2010secure,aliasgariBZS13}.

Given a set of clients $P$, each with some matrix $Z^{(j)}\in \mathbb{Z}_l^{n \times c}$, 
ordinary \texttt{SecureSum} computes $\sum_{j\in P}Z^{(j)}$ at the server in such a way 
that the server learns only the sum but nothing about the $Z^{(j)}$ matrices.
The main idea is as follows: using agreed upon random seeds clients jointly create random 
matrices $M^{(j)}\in \mathbb{Z}_l^{n \times c}$ with the property that $\sum_{j\in P}M^{(j)} = 0$.
Each client $j$ then obfuscates its data by computing $\tilde{Z}^{(j)} \coloneqq Z^{(j)} + M^{(j)}$ 
and sends this to the server.
From the perspective of the server each $\tilde{Z}^{(j)}$ is indistinguishable from uniformly 
random noise. However, when all parts are summed, the obfuscations cancel out and what remains 
is the desired answer: $\sum_{j\in P}\tilde{Z}^{(j)} = \sum_{j\in P}Z^{(j)}$. 
For technical details see \citet{BonawitzIKMMPRS17}.

For \method, we propose a modification of the above construction. 
Suppose we have a partition of the rows, $(R_j)_{j\in P}$, where each $R_j \subset [n]$. 
Each client $j$ knows its own $R_j$ and the server knows all $R_j$. 
\texttt{\oursum}'s task is to compute $Z \coloneqq\sum_{j\in P}Z^{(j)}$ in 
a distributed form in which each client $j\in P$ learns only the rows of $Z$ 
indexed by $R_j$, denoted $Z[R_j]$, and the server learns nothing.

For this, let $M^{(j)}$ and $\tilde{Z}^{(j)}$ be defined as above. 
In addition, let $\hat{Z}^{(j)}$ be equal to the obfuscated $\tilde{Z}^{(j)}$, 
except that the rows indexed by $R_j$ are completely set to 0.
Each client now instead sends $\hat{Z}^{(j)}$ to the server, which 
computes $\hat{Z} \coloneqq \sum_{j\in P}\hat{Z}^{(j)}$.
The server then redistributes $\hat{Z}$ among the clients, \ie each 
client $j$ receives $\hat{Z}[R_j]$.
Note that $\hat{Z}[R_j]=\sum_{k\in P\setminus\{j\}}\tilde{Z}^{(k)}[R_j]
= Z[R_j] - \tilde{Z}^{(j)}[R_j]$. 
Consequently, each client obtains the part of $Z$ corresponding to its own
data by computing $Z[R_j]=\hat{Z}[R_j] + \tilde{Z}^{(j)}[R_j]$.
By construction the shared quantities leak nothing to the server. 
Specifically $\hat{Z}^{(j)}$ is random noise with rows $R_j$ set to 0 
and $\hat{Z}$ is random noise since each block $\hat{Z}[R_j]$ remains 
obfuscated due to client $j$ not sending $M^{(j)}[R_j]$.

\texttt{\bfseries SecureHamming}\quad
Several cryptographic protocols for computing the Hamming distances between binary vectors exist. 
Here we propose two variants that are tailored to the setting of \method: one is based 
on the SHADE protocol~\citep{bringer2013shade}, which is easy to implement as it only 
requires only an \emph{oblivious transfer (OT)}~\citep{naor2001efficient} routine as 
cryptographic primitive. 
The other is based on \emph{partially homomorphic encryption (PHE)} and comes with 
lower communication cost. % see \Cref{app:crypto2}.

\emph{OT-based secure Hamming distance:}\quad
Let $b=(b_1,\dots,b_L)$ and $b'=(b'_1,\dots,b'_L)$ be the bit vectors, for which the Hamming 
distance should be computed, where $b$ is stored on a client $j$ and $b'$ on a client $k$. 

1) client $j$ creates $L$ random numbers $r_1,\dots,r_L$ uniformly in the range $[0,L-1]$. 
For each $l=1,\dots,L$, it then offers two values to be transferred to client $k$: $z^0_l=r_l+b_l$ or $z^1_l = r_l+\bar b_l$, for $\bar b_l = 1-b_l$ (here and in the following all calculations are performed in $\mathbb{Z}_L$, \ie in the integers \emph{modulo} $L$).

2) Client $k$ initiates an OT operation with input $b'_l$ and result $t_l$. 
That means, if $b'_l=0$ it will receive $t_l=z^0_l$ and if $b'_l=1$ it will receive $t_l=z^1_l$, 
but client $i$ will obtain no information which of the two cases occurred.
Note that in both cases, it holds that $t_l = r_l + b_l\oplus b'_l$. 
However, client $k$ gains no information about the value of $b_l$ from 
this, because of the uniformly random shift $r_l$ that is unknown to it. 

3) Clients $j$ and $k$ now compute $R=\sum_{l=1}^L r_l$ and $T=\sum_{l=1}^L t_l$, 
respectively, and send these values to the server. 

4) From $R$ and $T$, the server can infer the Hamming distance between $b$ and $b'$ as $T-R = \sum_{l=1}^L b_l\oplus b'_l = h(b,b')$. 

Performing these steps for all pairs of data points, the server obtains the Hamming matrix, $H\in\sN^{n\times n}$, but no other information about the data. The clients obtain no information about each others' data at all during the computation.
%

%However, it has sub-optimal communication cost because each Hamming distance computation is treated independently.

\emph{PHE-based secure Hamming distance:}
Homomorphic encryption is a framework that allows 
computing functions on encrypted arguments~\citep{acar2018survey}. 
The outcome is an encryption of the value that would have 
been the result of evaluating the target function on the 
plaintext arguments, but the computational devices gain 
no information about the actual plaintext. 
\emph{Fully homomorphic encryption(FHE)}, which allows 
arbitrary functions to be computed this way, is not 
efficient enough for practical usage so far~\citep{jiang2022fhebench}.
However, efficient implementation exist for 
the setting where cyphertexts only have to be 
added or subtracted, for example \emph{Paillier}'s~\citep{paillier1999public}. 
We exploit this paradigm of \emph{partially homomorphic encryption (PHE)} 
to compute the Hamming distances of a binary vector from 
clients $j$ with all binary vectors from a client $k$ in 
the following way: 

1) Client $j$ encrypts its own data vector $b^{(j)}\in\{0,1\}^L$, 
using its own public key and transfers the resulting vector 
$\boxed{y}:=\text{enc}(b^{(j)})$ (boxes indicate encrypted quantities)
to client $k$.
Because the data is encrypted, client $k$ can extract no 
information about client $j$'s data from this.

2) For any of its own data vectors $b^{(k)}\in\{0,1\}^L$,  
client $k$ creates a uniformly random value $r\in[0,L-1]$. 
It then computes the following function in a homomorphic 
way with encrypted input $y=\boxed{y}$ and plaintext input $x=b^{(k)}$:
$$f(y;x)= r + \sum_{l:x_l=1} y_l - \sum_{l:x_l=0} y_l $$
Because of the identity $h(x,y)=\sum_l[x_ly_l + \bar x_l\bar y_l]=\sum_{l:x_l=1} y_l + \sum_{l:x_l=0}\bar y_l
= \sum_{l:x_l=1} y_l - \sum_{l:x_l=0} y_l + n - \sum_l x_l$, the 
result, $\boxed{T}=f(\boxed{y};x)$, is the value $T = h(b^{(j)},b^{(k)}) + r - n + \sum_l b_l $ in encrypted form. 

4) Client $j$ send the value $R=r-n+\sum_l b_l$ to the server, 
and the value $\boxed{T}$ to client $j$. Client $j$ decrypts 
$\boxed{T}$ using its own secret key and sends the resulting 
value $T$ to the server. It gains no information about client
$k$'s data or the Hamming distance, because the added randomness 
gives $T$ a uniformly random distribution.

5) The server recovers $H(b^{(j)},b^{(k)})=T-R$ without gaining any other information about the clients' data. 

Note that for computing single Hamming distances, 
this protocol has no advantage over the OT-based 
variant. 
However, for computing all pairwise Hamming distances 
between the data of two clients, the PHE-based protocol
saves a factor $n^{(k)}$ in communication cost, because 
$\boxed{y}$ has to be transferred only once and can then 
be used repeatedly by client $k$ to compute the distances 
to all of its vectors.

\subsection{Analysis}\label{subsec:analysis}
In this section, we analyze the \emph{efficacy}, \emph{privacy},
\emph{efficiency} and \emph{robustness} of \Cref{alg:crossclientLP}. 

\myparagraph{Efficacy}\quad \Cref{alg:crossclientLP} 
performs label propagation along the data graph, 
as the classical LP algorithm~\citep{zhu2005semi} 
does when data is centralized. 
The similarity measure used is cosine similarity 
estimated via the Hamming distance of the LSH binary vectors.
How close this is to the actual cosine similarity is 
determined by $L$, the LSH vector length. 
In practice, we observe no difference in behavior between
them already for reasonably small values, \eg $L=4096$.

% \myparagraph{Privacy }\quad The main insight is that \Cref{alg:crossclientLP} 
% adheres to the federated learning principle that clients 
% do not have to share their data or labels with any other party.
% %
% This is ensured by the fact that all \emph{cross-steps} are 
% computed using cryptographically secure methods.
% %
% The only information seen by the server about client data is contained in 
% the matrix of Hamming distances $H$, from which it can approximately recover 
% the matrix of cosine similarities $W$.
% %
% While certainly influenced by the client data, we consider $W$ (and therefore $H$) 
% a rather benign object for a non-hostile server to have access to because cosine similarity depends only on angles, 
% hence any rescaling and rotation of the client input vectors would result in the same $W$ matrix. 
% %
% Clients do not see $W$ but they see some of the columns of $S$. These 
% reflect how their labeled data can influence all other data points according 
% to the data graph as estimated from the participating clients' vectors. 
% %
% However, this influence is unnormalized and hence the influence relative to other clients cannot be known.

\myparagraph{Privacy }\quad The main insight is that \Cref{alg:crossclientLP} 
adheres to the federated learning principle that clients 
do not have to share their data or labels with any other party.
This is ensured by the fact that all \emph{cross-steps} are 
computed using cryptographically secure methods.
There are two potential places where clients share some information relating to their data.
The first is the matrix of Hamming distances $H$ that is sent to the server, and from which the server can approximately recover 
the matrix of cosine similarities $W$.
While certainly influenced by the client data, we consider $W$ (and therefore $H$) 
a rather benign object for a non-hostile server to have access to because cosine similarity depends only on angles, 
hence any rescaling and rotation of the client input vectors would result in the same $W$ matrix. 
A second source of information sharing is during phase $3$ where each client receives the columns of $S$ that correspond to their data. 
Such columns reflect how their labeled data can influence all other data points according 
to the data graph as estimated from the participating clients' vectors. 
However, we stress that this influence is unnormalized and hence the influence relative to other clients cannot be known.
%
% Additionally, we believe any leakage of this form is integral to any algorithm than performs meaningful training (as opposed to say an algorithm that returns true randomness and thus would not leak any information but is effectively useless).

\myparagraph{Computational Complexity}\quad 
The computational cost of \method is determined by factors: the cryptographic subroutines and the numeric operation. 
The contribution of the former depends heavily on the underlying implementation and available hardware support, so we do not discuss it here. 
The cost of the numeric operations can be derived in explicit form. Assume that the number of clients per 
batch is $p$. Each client has $n$ data points in total, out of which $m$ are labeled (for simplicity, we assume
$n$ and $m$ to be identical across clients here). Let the feature dimensionality be $d$ and the number of classes $C$. 
Then, to run XCLP with $L$-dimensional bit vectors, each client computes an $n\times L$ binary data matrix, which has complexity $O(dnL)$. 
Then, the clients jointly compute all $p^2n^2/2$ pairwise Hamming distances, which requires $O(Lp^2n^2)$ operations and has overall 
complexity $O(Lpn^2)$ if run in parallel by the clients. The server inverts the matrix at cost $O(p^3n^3)$. 
Then, each client multiplies an $pn\times m$ sized part of the resulting matrix with their $m\times C$ label matrix, 
which costs $O(Cpnm)$ per client, and also has overall complexity $O(Cpnm)$ if run on the clients in parallel. 
Overall, the most costly step is the matrix inversion, but that is done on the server, which we assume to have much higher compute capabilities.
Assuming $L>C$ and $np>d$, the clients’ costs are dominated by the $O(Lpn^2)$ term. 

For comparison, to run per-client LP, each client has to compute $n^2/2$ $d$-dimensional inner products, 
invert the resulting $n\times n$ matrix, and multiply a sub-matrix of size $m\times n$ it with a label 
matrix of size $m \times C$. The complexity per client is $O(dn^2+n^3+Cmn)$, which is also the total complexity, 
if all clients can operate in parallel. 
In practice, we expect $d>n>C$, so the dominant term is $O(dn^2)$. The server does not contribute any computation. 
Consequently, with a typical trade-off of, e.g. $L=8d$, XCLP requires $8p$ times more numeric operations than local LP. 
Note, however, that one has good control over the total cost, as $L$ and $p$ (and potentially $d$) are design choices. 

On an absolute scale,, all of these values are rather small. For example, with $p=10$, $d=512$, and $L=4096$, 
the number of numeric operations a client has to perform per datapoint is in the order of $10^7$. In many settings,
this can be expected to be less than the the operations needed to compute the datapoint’s feature representation in
the first place, \eg when using a (even small) neural network for that purpose. 

\myparagraph{Communication Efficiency}\quad \method incurs 
communication costs at two steps of \Cref{alg:crossclientLP}.
With the OT-based protocol for computing the 
$n^{(j)}\times n^{(k)}$ Hamming matrix 
between two clients $j$ and $k$, client $j$ sends 
$n^{(j)}n^{(k)}L$ integer values in 
$\mathbb{Z}_L$ to client $k$.
With the enhanced PHE-based protocol, this amount is 
reduced to sending $n^{(j)}L$ encrypted values from 
$j$ to $k$ and $n^{(j)}n^{(k)}$ in the opposite direction.
Each of the two clients sends $n^{(j)}n^{(k)}$ integer 
values in $\mathbb{Z}_L$ to the server. 
%, and $pn^{(k)}$ integers to the server.
%
To propagate the labels via the distributed matrix multiplication, 
each client $j$ first receives from the server a matrix of size 
$n\times l^{(j)}$.
It transmits a matrix of size $n\times C$ to the server, 
and receives a matrix of size $n^{(j)}\times C$ 
back from it. 
In particular, \method requires only a constant number of communication 
steps, which is in contrast to other methods that train iteratively.

\myparagraph{Robustness}\quad 
In \emph{cross-device} FL clients may be unreliable and prone to disconnecting
spontaneously. Therefore, it is important that FL algorithms can still 
execute even in the event of intermediate client dropouts.  
This is indeed the case for \Cref{alg:crossclientLP}:
a client dropping out before the \texttt{SecureHamming} step 
(line \ref{xcLPline:secure_hamming}), is equivalent to it 
not having been in $P$ in the first place.
Since the Hamming computation is executed pairwise, a client dropping 
out during this step has no effect on the computation of other clients. 
The result will be missing entries in the matrix, $H$, which 
the server can remove, thereby leading to the same outcome as if the 
client had dropped out earlier.
If clients drop out after $H$ has been computed, but before the 
\texttt{\oursum} step (line \ref{xcLPline:secure_sum}), they will 
have contributed to the estimate of the data graph, but they will 
not contribute label information to the propagation step. This has 
the same effect as if the client only had unlabeled data. 
If clients drop out within \texttt{\oursum}, after the obfuscation 
matrices have been agreed on but before the server has computed 
$\hat Z$, then the secure summation could not be completed. 
To recover, the server can simply restart the \texttt{\oursum} 
step without the dropped client. 
Any later dropout will only result in that client not receiving 
labels for its data, but it will not affect the results for 
the other clients.

\begin{table*}[t]
\caption{\method for prediction (Fed-ISIC2019 dataset): Classification accuracy [in \%] with 
two different preprocessing functions (pretrained and finetuned) and training sets of different size (average and standard deviation across three runs).
}
\label{tab:fedisic_results2}
\begin{center}
\small
\begin{tabular}{|c||c|c||c|c|c|}\hline
 & \multicolumn{2}{c||}{pretrained} & \multicolumn{3}{c|}{finetuned} \\
\textit{training set size} & \emph{per-client LP} & \emph{\method} & \emph{FedAvg} & \emph{per-client LP} & \emph{\method} \\
 \hline\hline
 $n=954$ & $30.70 \pm 1.25$ & $\bf34.00 \pm 1.12$ & $43.78 \pm 0.88$ & $45.35 \pm 1.25$ & $\bf47.48 \pm 1.12$
 \\\hline
 $n=1882$ & $31.40 \pm 1.18$ & $\bf34.61 \pm 0.93$ & $46.83 \pm 0.75$ & $48.10 \pm 1.18$ & $\bf52.34 \pm 0.93$
 \\\hline
 $n=3744$ & $37.41 \pm 0.20$ & $\bf38.49 \pm 0.88$ & $52.87 \pm 0.14$ & $56.98 \pm 0.20$ & $\bf59.49 \pm 0.88$
 \\\hline
 $n=18597$ & $\bf55.10 \pm 0.51$ & $53.38 \pm 0.67$& $63.78 \pm 0.20$ & $73.30 \pm 0.51$ & $\bf74.22 \pm 0.67$
\\\hline
\end{tabular}
\end{center}
\end{table*}

\begin{table}[t]
\centering
\caption{\method for prediction (Fed-ISIC2019 dataset): Classification accuracy [in \%]
for leave-one-client-out experiments (average and standard deviation across three runs).}
\label{tab:fedisic_results_clients}
\small
\begin{tabular}{ |c|c|c| }
\hline
\textit{left-out client} & \textit{FedAvg} & \textit{\method} \\
 \hline\hline
Client 1 & $35.15 \pm 0.96$ & $\bf38.89 \pm 1.34$ \\
\hline
Client 2 & $\bf68.75 \pm 0.72$ & $67.05 \pm 0.54$\\
\hline
Client 3 & $51.94 \pm 2.52$ & $\bf62.02 \pm 1.13$ \\
\hline
Client 4 & $42.72 \pm 0.79$ & $\bf54.84 \pm 0.78$ \\
\hline
Client 5 & $41.53 \pm 1.60$ & $\bf52.28 \pm 0.58$ \\
\hline
Client 6 & $46.92 \pm 3.69$ & $\bf61.27 \pm 3.19$ \\
 \hline
\end{tabular}
\end{table}
\begin{table}[t]
\caption{\method for prediction (Fed-ISIC2019 dataset): Classification accuracy [in \%] when different fractions, $\alpha$, of training data are labeled. {FedAvg} and \emph{\method(labeled)} use only the labeled part of the training set, \emph{\method(labeled+unlabeled)} uses also the unlabeled part.}
\label{tab:fedisic_results}
\begin{center}
\small
% \begin{center}
\begin{tabular}{ |c|c|c|c| }
\hline
 & \emph{FedAvg} & \emph{\method(labeled)} & \emph{\method(labeled+unlabeled)} \\
 \hline
 $\alpha=0.05$ & $43.78 \pm 0.88$ & $47.48 \pm 1.12$ & $\bf49.50 \pm 1.42$ \\
 \hline
 $\alpha=0.1$ & $46.83 \pm 0.75$ & $52.34 \pm 0.93$ & $\bf53.78 \pm 0.90$ \\
 \hline
 $\alpha=0.2$ & $52.87 \pm 0.14$ & $59.49 \pm 0.88$ & $\bf61.09 \pm 0.66$ \\
 \hline
 $\alpha=1.0$ & $63.78 \pm 0.20$ & $\bf74.22 \pm 0.67$ & $\bf74.22 \pm 0.67$\\
 \hline
\end{tabular}
\end{center}
\end{table}

\section{Experiments}\label{sec:experiments}

In the following section we report experimental results for \method. 
We present two applications in the context of federated learning. 
In \Cref{sec:LPpred} we illustrate how \method can be used in a 
one-shot way to infer labels at prediction time. 
In \Cref{sec:FedProp} we show how \method can be used for federated 
semi-supervised learning by integrating it into a federated averaging 
training loop.
As our emphasis here is on accuracy, not real-world efficiency, we 
use a simulated setting of federated learning, rather than physically 
distributing the clients across multiple devices.  
Therefore, we also use plaintext placeholders for the cryptographic 
steps that have identical output.
Source code for our experiments can be found at https://github.com/jonnyascott/xclp.

\subsection{\method for prediction}\label{sec:LPpred}

The most straightforward application of \method is as a method to predict labels for new data at inference time. 
For this setting suppose a set of clients, $P$, possess training data $X^{(j)}\in\mathcal{X}^{n^{(j)}}$ from some input space $\mathcal{X}$, 
and a (potentially partial) label matrix  $Y^{(j)}\in\{0,1\}^{n^{(j)}\times C}$. 
The goal is to infer labels for new batches of data, $X^{(j)}_{\text{new}}$. 
Note that the above setting is general enough to encompass a number of settings, 
including clients with fully labeled or fully unlabeled training data. 
Also included is the possibility that a client $j$ has no training data to 
contribute, $X^{(j)}=\emptyset$, but has a batch of new data to be labeled, $X^{(j)}_{\text{new}}\neq\emptyset$.

By $h : \mathcal{X}\rightarrow \R^d$ we denote a preprocessing function, such as a feature extractor. 
Each client applies $h$ to all their data points, train and new, to obtain their input vectors 
to the \method routine, $V^{(j)} \coloneqq h\big( X^{(j)} \cup X^{(j)}_{\text{new}} \big)$.
Running \Cref{alg:crossclientLP} on $(V^{(j)}, Y^{(j)})_{j \in P}$, each client obtains $\hat{Y}^{(j)}$, which are label assignments for all of their data points, in particular including $X^{(j)}_{\text{new}}$ as desired. 

\myparagraph{Experimental Setup}\quad
We use the Fed-ISIC2019 dataset~\citep{flamby}, a real-world benchmark for federated 
classification of medical images. It consists of a total of 23247 images across 6 clients.
The dataset is highly heterogeneous in terms of the amount of data per client, 
the classes present at each client as well as visual content of the images.  
As baseline classifier, we follow \citep{flamby} and use an EfficientNet~\citep{efficientnet}, 
pretrained on ImageNet, which we finetune using federated averaging.
As preprocessing functions, $h$, we use the feature extraction layers of 
the network either at the point of initialization (pretrained) or after the finetuning.
\Cref{app:experiments1} gives full details of the experimental setup.

\myparagraph{Results}\quad
\Cref{tab:fedisic_results2}\ reports results for the setting in 
which all clients contribute fully-labeled training data and have 
new data that should be classified.
The left columns (``pretrained") illustrate the one-shot setting: 
no network training is required, only features are extracted once 
using a pretrained network, and \method is run once to infer labels.
\method performs better than per-client label propagation here, except 
for the largest dataset size, indicating that in the limited data regime,
it is indeed beneficial to build the data graph jointly from all data rather
than separately on each clients.
The other three columns (``finetuned") illustrate that with a better 
--task-adapted-- feature representation, \method still outperforms 
per-client LP, and also achieves better accuracy than predicting
labels using only the network.

\Cref{tab:fedisic_results_clients} reports results for a more challenging 
setting. We adopt a leave-one-client-out setup in which one client
does not contribute labeled training but instead its data is meant to be 
classified.
Given the heterogeneity of the clients, this means the classifiers have to
overcome a substantial distribution shift. 
\method achieves better results than a network trained by federated 
averaging in all but one case, where in several cases the advantage
is quite substantial. 
Note that per-client LP is not applicable here, as the new data 
is all located on a client that does not have labeled training 
data.

Finally, we also conduct on ablation study on the effect of unlabeled training
data on \method, that is when $Y^{(j)}$ is a (strictly) partial label matrix.
In this case the unlabeled training data does not contribute label information
towards inference on $X^{(j)}_{\text{new}}$, but does contribute to a more densely
sampled graph. 
%
%The results are shown in \Cref{app:ablation_unlabeled}.
%
\Cref{tab:fedisic_results} shows the results: the \emph{FedAvg} column is 
identical to the one in \Cref{tab:fedisic_results2}, because the federated 
averaging  training does not benefit from the additionally available unlabeled 
training data. 
Similarly, the result for \emph{\method} with only labeled data are the 
same as for \method in \Cref{tab:fedisic_results2}.
However, allowing \method to exploit the additional unlabeled data, 
however, indeed improves the accuracy further. 
This results once again shows the benefits of exploiting unlabeled data,
especially when the amount of label data is small.

%One can see that additional unlabeled data always improves \method's performance. This supports the hypothesis that a better estimation of the data graph, as enabled by having more unlabeled data, positively affects the classification accuracy, even if the number of labeled data points remains unchanged. 

\SetNlSty{}{}{}
\IncMargin{-2em}
\begin{algorithm}[t]
\caption{\Fedprop}\label{alg:fedprop}
\Indm  
\Input{partially labeled training data $(X^{(j)}, Y^{(j)})_{j=1}^m$}%\tcp*{stored on-device at clients}}
\Indp
$\theta=(\phi,\psi) \gets \texttt{InitializeModelParameters}$\label{algline:rand_init}\!\!\!\!\!\;
%$\theta \gets \texttt{FederatedOptimization}(X^{(j)}_L, y^{(j)}_L)$\label{algline:init_train} %\tcp*{initial training on labeled examples}
\For{round $t \in [1, \dots T]$}{
%\COMMENT{Second phase training on all examples}
$P \gets \text{server randomly selects } \tau m \text{ clients}$\;
Server broadcasts $\theta$ to each client in $P$ \label{algline:broadcast}\;
\For{client $j \in P$ in parallel}{
$V^{(j)} \gets f_{\phi}(X^{(j)})$\; \label{algline:features}%\tcp*{embed labeled and unlabeled data}
$\hat{y}^{(j)}, \omega^{(j)} \gets \texttt{\method}(V^{(j)}, Y^{(j)}, P, \text{ Server})$ \label{algline:cross_client_LP}\; %\tcp*{compute pseudo-labels and confidences} 
\mbox{$\theta^{(j)} \gets \texttt{ClientUpdate}(X^{(j)}, \hat{y}^{(j)}, \omega^{(j)}; \theta)$\!\!\!\!\!\!\!\!\!\!\!\!\!\!\!} \label{algline:localtrain}\;
Client $j$ sends $\theta^{(j)}$ to the server \label{algline:client_sends}\;
}
$\theta \gets \texttt{ServerUpdate}\big( (\theta^{(j)})_{j\in P})\big)$ \label{algline:server_avgs}
}
\Indm  
\Output{model parameters $\theta$}
\Indp
\end{algorithm}

\subsection{\method for federated semi-supervised learning}\label{sec:FedProp}

We now describe how \method can be applied iteratively during a 
training loop in the context of federated semi-supervised learning.
\emph{\Fedprop}, shown in pseudocode in \Cref{alg:fedprop}, follows 
a general FL template of alternating local and global model updates. 
As such, it is compatible with most existing FL optimization schemes, 
such as \emph{\mbox{FedAvg}}~\citep{fedavg}, \emph{FedProx}~\citep{fedprox}, 
or \emph{SCAFFOLD}~\citep{scaffold}. The choice of scheme determines 
the exact form of the \texttt{ClientUpdate} and \texttt{ServerUpdate} routines.

The first step (line \ref{algline:rand_init}) is to initialize the model parameters, $\theta=(\phi,\psi)$,
where $f_{\phi}:\mathcal{X}\to\R^d$ is the feature extraction part of a neural network and $f_{\psi}:\R^d\to\R^C$ is the classifier head.
The initialization could be random, using weights of a pretrained network, 
by an unsupervised technique, such as contrastive learning, or by a 
supervised step, such as federated training on only the labeled examples.

We then iterate the main training loop over $T$ rounds. 
To start each round the server samples some fraction $\tau$ 
of the $m$ total clients. 
These clients receive the current model parameters from the server 
(line \ref{algline:broadcast})
and embed their labeled and unlabeled data with the feature extractor, $f_\phi$ (line \ref{algline:features}).
Clients and server then collaboratively run \method on these feature vectors (line \ref{algline:cross_client_LP}).
As output of this step each client updates the pseudo-labels and confidence 
values for their unlabeled data, which they then use for local supervised training (line \ref{algline:localtrain}).
Lastly, clients send the updated local models to the server 
(line \ref{algline:client_sends}) which aggregates them 
(line \ref{algline:server_avgs}).

The motivation for this approach comes from the insight gained in \Cref{sec:LPpred}, that \method assigns high
quality labels when run on features obtained from a trained network. Crucially, pseudo-labels assigned
by \method are always recomputed when a client is sampled. Thus as the network features improve so too does the quality 
of the pseudo-labeling.

\myparagraph{Experimental Setup}\quad
We evaluate the accuracy of \mbox{\Fedprop} against other methods 
for federated SSL as well as report on ablation studies.
We adopt a standard federated averaging scheme, in which 
\texttt{ClientUpdate} consists of running $5$ epochs 
of SGD with confidence-weighted cross-entropy loss on the local device and \texttt{ServerUpdate} simply 
averages the local client models. 

We use three standard datasets: CIFAR-10 \citep{cifar}, which has 10 classes 
and is used in several previous federated SSL works~\citep{FedMatch,fedsiam,fedsem}, 
as well as the more difficult CIFAR-100 \citep{cifar} and Mini-ImageNet
\citep{miniimagenet} which have 100 classes. 
To the best of our knowledge ours is the first work in this 
federated SSL setting to evaluate on these more challenging 
datasets. 
The datasets are split in different ways (different number of 
clients, different amounts of labeled data, \iid vs non-\iid) 
to simulate a diverse range of federated settings.

\begin{table*}[t]
\centering
\caption{\method for federated SSL: classification accuracy [in \%] on federated CIFAR-10. $m$ is the number of clients, $m_L$ the number of clients with labeled data, $n_L$ is the total number of labels across all clients.
\iid and non-\iid refer to how the data is split among the clients. For details, see the main text and~\Cref{app:experiments2}.
}
\label{tab:cifar10-results}
\small
\begin{tabular}{ |c|c|c|c|c| }
\hline
\multicolumn{5}{|c|}{\textbf{CIFAR-10, \iid} ($m=100$)} \\
\hline
 & \multicolumn{2}{c|}{$m_L=100$} & \multicolumn{2}{c|}{$m_L=50$} \\
 \hline
 \textbf{Method} & $n_L=1000$ & $n_L=5000$ & $n_L=1000$ & $n_L=5000$\\
 \hline
 FedAvg (labeled only) & $55.46 \pm 0.43$& $76.13 \pm 0.46$& $56.97 \pm 0.59$ & $80.36 \pm 0.07$ \\
 \hline
 FedAvg+perclientLP &$61.75\pm 2.22$ & $85.11 \pm 0.73$ & $65.29 \pm 2.50$ & $84.41 \pm 0.25$\\
 FedAvg+network & $60.12 \pm 0.15$ & $79.45\pm0.31$& $59.14\pm 0.35$ & $81.04\pm0.20$ \\
 FedMatch & $50.93\pm 0.56$ & $72.22\pm0.14$& $57.10\pm0.46$ & $77.80\pm0.32$ \\
 FedSiam & $67.02\pm0.98$ & $82.06\pm0.56 $& $62.98\pm1.61$ & $78.45\pm0.34$ \\
 FedSem+ & $59.98\pm 0.49$ & $79.49\pm0.15$& $59.67\pm 0.47$& $80.94\pm 0.25$\\
 % FedAvg+MT &$62.37\pm 1.69$ & $84.92 \pm 0.64$ & $70.14 \pm 1.87$ & $85.34 \pm 0.18$ \\
 % \hline
 \Fedprop (ours) & $\bf 70.91\pm0.71$ & $\bf 86.65 \pm 0.16$ & $\bf 70.81\pm 1.65$ & $\bf 86.29\pm 0.34$ \\
 % \methodMTup (ours) & $\bf72.58\pm 0.36$  & $\bf88.17\pm 0.18$& $\bf73.63\pm 1.99$  & $\bf87.54\pm 0.14$ \\
 \hline
 \hline
 \multicolumn{5}{|c|}{\textbf{CIFAR-10, non-\iid} ($m=100$)} \\
\hline
 & \multicolumn{2}{c|}{$m_L=100$} & \multicolumn{2}{c|}{$m_L=50$} \\
 \hline
 \textbf{Method} & $n_L=1000$ & $n_L=5000$ & $n_L=1000$ & $n_L=5000$\\
 \hline
 FedAvg (labeled only) & $50.94 \pm 0.14$& $75.34 \pm 1.38$& $53.26 \pm 0.69$ & $79.65\pm 0.12$ \\
\hline
 FedAvg+perclientLP & $50.94 \pm 0.14$ & $76.61\pm 1.50$ & $53.26 \pm 0.69$ & $79.65\pm 0.12$ \\
 FedAvg+network & $60.60\pm0.60$ & $80.07\pm 0.53$ & $59.82\pm1.05$ & $81.14\pm0.23$\\
 FedMatch & $50.71\pm1.57$ & $71.99\pm0.70$ & $48.24\pm0.86$ & $66.37\pm 0.41$ \\
 FedSiam & $67.85\pm 0.26$ & $82.23\pm0.46$& $62.29\pm1.84$ & $78.84\pm0.72$  \\
 FedSem+ & $60.93\pm0.97$ & $79.70 \pm 0.78$& $59.74\pm 0.74$ & $81.30 \pm0.09$ \\
% \hline
 \Fedprop (ours)& $\bf73.76\pm0.71$ & $\bf 85.53\pm 0.56$ & $\bf70.01\pm 1.29$ & $\bf 85.42\pm 0.43$ \\
 \hline
\end{tabular}
\end{table*}
\begin{table*}[t]
\caption{\method for federated SSL: classification accuracy [in \%] on federated CIFAR-100 and Mini-ImageNet. $m$ is the number of clients, $m_L$ the number of clients with labeled data, $n_L$ is the total number of labels across all clients.}
\label{tab:joint-results}
\centering
\small
\begin{tabular}{ |c|c|c|c|c| }
\hline
& \multicolumn{2}{|c|}{\textbf{CIFAR-100, \iid}} & \multicolumn{2}{|c|}{\textbf{Mini-ImageNet, \iid}} \\
\hline
 & $m=m_L=50$  & $m=m_L=100$ & $m=m_L=50$  & $m=m_L=100$ \\
 \hline
 \textbf{Method} & $n_L = 5000$ & $n_L=10000$ & $n_L = 5000$ & $n_L=10000$ \\
 \hline
 FedAvg (labeled only) & $43.80 \pm 0.19 $& $53.91 \pm 0.25 $ & $23.39 \pm 0.52$& $31.72 \pm 0.54 $ \\
  \hline
 FedAvg+network & $43.80 \pm 0.19 $ & $54.19\pm 0.21$& $23.98\pm 0.36$& $31.86\pm 0.57$ \\
 FedAvg+perclientLP & $43.82 \pm 0.59 $ & $54.38 \pm 0.36 $ & $25.53\pm 0.22$ & $33.09\pm 0.62$ \\
 % FedAvg+MT & $49.09 \pm 0.38$ & $56.05\pm 0.23$ & $25.98\pm 0.70$ & $33.20\pm 0.68$ \\
  % \hline
 \Fedprop (ours) &$\bf 50.19\pm 0.60$ & $\bf 57.00 \pm 0.08$ & $\bf26.93\pm 0.41 $ & $\bf35.78\pm 0.56 $\\
% \methodMTup (ours) &$\bf 50.60 \pm0.28$& $\bf 58.66\pm 0.28$ & $25.62\pm 0.75$ & $33.24\pm 0.71$  \\
 \hline
\end{tabular}
\end{table*}

We compare \Fedprop to a broad range of other methods.
To enable a fair comparison of results, all methods use 
the same network architecture, and hyper-parameters are 
chosen individually to maximize each method's performance.
From the existing federated SSL literature, we report results 
for \emph{FedMatch} \citep{FedMatch}, \emph{FedSiam} \citep{fedsiam}
and \emph{FedSem+}, which follows \citep{fedsem} but 
additionally uses confidence-based sample weights, as 
we found these to consistently improve its accuracy.
Additional baselines are two methods that follow the same 
structure as \Cref{alg:fedprop} but use alternative ways 
to obtain pseudo-labels: from per-client label propagation 
(FedAvg+perclientLP) or from the network's classifier 
predictions (FedAvg+network). 
Finally, we also report results for training in a 
supervised manner on only the available labeled 
examples, \emph{FedAvg (labeled only)}. 
Note that we do not include comparisons to \citet{grad_diversity} and \citet{semiFL} as these methods address a different 
federated SSL scenario in which the server has access to labeled data while the clients have no labels.

\myparagraph{Results}\quad We report the results of our experiments in Tables~\ref{tab:cifar10-results} and \ref{tab:joint-results} 
as the average accuracy and standard deviation over three random splits of the data for each setting. 
\Cref{tab:cifar10-results} provides a comparison of \method to other approaches and baselines in the standard 
setting of CIFAR-10 with 100 clients, which has been used in prior work.
In each case, we report results when 1{,}000 or 5{,}000 of the data points are labeled.
Either all or half of the clients have labels, with classes distributed either \iid or non-\iid across clients. 
\Cref{tab:joint-results} reports on the harder situation with many more classes, which prior work has not attempted.
Across the board, \method achieves the best results among all methods,
while of the other methods, none has a consistent advantage over the others.
In addition to these general observations the results offer a number of more specific insights.

Firstly, in nearly all cases semi-supervised methods outperformed the labeled only FedAvg baseline. This is to be expected given the additional (unlabeled) data available to the SSL methods. A notable exception to this, however, is FedAvg + perclientLP in the non-\iid scenario. In this case in three of the four cases perclient-LP actually degraded the performance of the initial supervised model, which caused automatic model selection to deactivate it. A likely reason for this poor performance is the presence of classes in the client test data which do not appear in the labeled portion of the clients training data. In this situation perclient-LP is not able to predict for these classes on the test data. In contrast \method is unlikely to be affected by this issue, as more clients are pooled together and hence more classes are present in the labeled training data. This is reflected in the strong performance of FedAvg + \method in the non-\iid setting.

Secondly, we observe that the biggest gains from incorporating \method occur in the regime where little labeled data is available. In particular we see that with ample labeled data available per client, in the \iid settings, perclientLP performs not so much worse than \method. This is to be expected as when each client already has ample labels available then the potential gain of collaboration with other clients is of course lower.

\section{Conclusions and Limitations}
In this work we introduced \method, a method for transductively predicting 
labels for unlabeled data points in a federated setting, where
the data is distributed across multiple clients.
It makes use of cryptographic routines to preserve data confidentiality 
when estimating a joint neighborhood graph over the data and 
propagates label information across this graph by distributing 
the computation among the clients. 
We presented two applications of \method, inferring labels for new (test) 
data, and training on partly labeled data in a federated semi-supervised 
setting.
In our experiments \method led to substantial improvements in 
classification accuracy in both applications, especially in the
most challenging (but often realistic) setting when the amount 
of labeled data per client is limited.

\method ensures that a client's data remains anonymous, in the 
sense that it is not directly exposed to or shared with any other 
party, a notion of privacy we referred to as data confidentiality. This is achieved through the guarantees provided by the cyryptographic subroutines.  
While data confidentiality is a fundamental building block of private machine learning paradigms, such as 
federated learning, on its own it does not guarantee
that nothing can be learned about the clients or their data.
Indeed it is common practice in private machine learning
to combine data confidentiality with other notions of 
privacy, such as differential privacy, in order to obtain
stronger privacy guarantees.
Extending \method to integrate such notions of privacy is an interesting direction for future work.

\bibliography{ms}
\bibliographystyle{tmlr}

%%%%%%%%%%%%%%%%%%%%%%%%%%%%%%%%%%%%%%%%%%%%%%%%%%%%%%%%%%%%%%%%%%%%%%%%%%%%%%%
%%%%%%%%%%%%%%%%%%%%%%%%%%%%%%%%%%%%%%%%%%%%%%%%%%%%%%%%%%%%%%%%%%%%%%%%%%%%%%%
% APPENDIX
%%%%%%%%%%%%%%%%%%%%%%%%%%%%%%%%%%%%%%%%%%%%%%%%%%%%%%%%%%%%%%%%%%%%%%%%%%%%%%%
%%%%%%%%%%%%%%%%%%%%%%%%%%%%%%%%%%%%%%%%%%%%%%%%%%%%%%%%%%%%%%%%%%%%%%%%%%%%%%%
\newpage

\appendix
\onecolumn

\section{Experimental Details}

\subsection{\method for Prediction} \label{app:experiments1}

\myparagraph{Dataset}\quad We use the Fed-ISIC2019 \citep{flamby} dataset which contains over 20, 000 images of skin lesions. The task is to predict melanoma (cancer) types. There are 6 clients, naturally defined by hospital and scanner used. As a result the data of each client is highly heterogeneous in terms of the amount of data per client (12413, 3954, 3363, 2259,
819, 439 examples for each client respectively), the classes present at each client as well as visual content of the images. Due to the class imbalance in the dataset the evaluation metric used is balanced accuracy.

\myparagraph{Network}\quad Following \cite{flamby} we use an EfficientNet \citep{efficientnet} pretrained on ImageNet, which we denote by $f$. We initialize a new final linear layer and fine-tune the whole network using federated averaging as described in \cite{flamby}.

\myparagraph{Hyper-parameters}\quad We set all hyper-parameters for \texttt{FederatedAveraging} to the values specified in \cite{flamby} except we increase the number of training rounds to $T=40$ as we found that the accuracy to improve with further training. Parameters for \method (LSH dimension, $k$-NN parameter) are chosen using cross-validation. We use $L=1024$ and $k=3$. We fix the parameter $\alpha = 0.99$.

\subsection{\method for federated semi-supervised learning} \label{app:experiments2}

\myparagraph{Datasets}\quad We evaluate \method on three standard datasets for 
multi-class classification: CIFAR-10 \citep{cifar}, which has 10 classes 
and is used in previous federated SSL works, 
as well as the more difficult CIFAR-100 \citep{cifar} and Mini-ImageNet
\citep{miniimagenet} which both have 100 classes. To the best of our 
knowledge ours is the first work in this federated SSL setting to 
evaluate on these more challenging datasets. 
All three datasets consist of $60{,}000$ images which we split into training sets of size $n \coloneqq 50{,}000$ and test sets of size $10{,}000$.
From the training set, $n_L$ examples are labeled and the remaining $n - n_L$ are unlabeled. 
For CIFAR-10 we evaluate with $n_L = 1{,}000 \text{ and } 5{,}000$. For CIFAR-100 and Mini-ImageNet we take $n_L = 5{,}000$ and $10{,}000$.

\myparagraph{Federated Setup}\quad We simulate a FL scenario by splitting the training data (labeled and unlabeled) between $m$ clients. $m_L$ of these have partly labeled data, while the others have only unlabeled data. 
Each client is assigned a total of $n/m$ data points of which $n_L / m_L$ are labeled if the client is one of the $m_L$ which possess labels. We simulate statistical heterogeneity among the clients by controlling the number of classes each client has access to. In the \iid setting all clients have uniform class distributions and receive an equal number of labels of each class. In the non-\iid setting we assign a class distribution to each client and clients receive labels according to their own distribution.

\myparagraph{Networks}\quad Following prior work, we use 13-layer CNNs~\citep{meanteacher} for CIFAR-10 and 100 and a ResNet-18 \citep{resnet18}
for Mini-ImageNet. Feature extractors are all layers except the last fully connected one, thus embeddings have dimension 128 and 512, respectively.

\myparagraph{Hyper-parameters}\quad We choose hyper-parameters for all methods based on training progress 
(LSH dimension, $k$-NN parameter) or accuracy on a held-out validation set consisting of 10\% 
of the training data (batch size, learning rate). %, or keep the default values. 
%
%Weight decay and data augmentation were kept at the values reported in~\citet{meanteacher}.  

\myparagraph{Federated learning parameters}\quad We set the number of clients to $m=100$, except for our experiments on CIFAR-100 and Mini-ImageNet with $n_L=5000$. In these cases we set $m=50$ as it is not possible to create an \iid split of the data over 100 clients since the number of classes (C=100) is too large. For CIFAR-10 we set the number of clients which possess labels to $m_L=100$ and $m_L = 50$. On CIFAR-100 and Mini-ImageNet we set $m_L = m$. 

The \texttt{ClientUpdate} step corresponds to $E$ epochs of stochastic gradient descent (SGD) of a loss function. We set the number local epochs to $E=5$ and the loss function is (per sample weighted) cross-entropy loss. The \texttt{ServerUpdate} step corresponds to averaging the model updates: \[\texttt{ServerUpdate}(\theta^{(j)} \text{ for } j\in P) = \frac{1}{|P|}\sum_{j\in P}\theta^{(j)}.\]

The number of training rounds is set to $T=1500$ and the number of clients sampled by the server per training round is set to $5$, so $\tau = 0.05$ when $m=100$ and $\tau = 0.1$ when $m=50$. Note that when $m_L < m$ we ensure that the server samples $\tau m_L$ clients from the labeled portion (and $\tau (m-m_L)$ from the unlabeled) to ensure that there are some labels present in the graph.

\myparagraph{Network training parameters } We use standard data augmentation following \cite{meanteacher}. On CIFAR-10 and CIFAR-100 this is performed by 4×4 random translations followed by a random horizontal flip. On Mini-ImageNet, each image is randomly
rotated by 10 degrees before a random horizontal flip. We use weight decay for all network parameters which is set to $2\times 10^{-4}$. When carrying out SGD in the \texttt{ClientUpdate} we use batches of data $B = B_L \cup B_U$ where $B_L$ is a batch of labeled data and $B_U$ is a batch of pseudo-labeled (previously unlabeled) data. We set $|B_L|$ according to how many labeled samples the client has available, $|B_L| = \min(50, \#labels)$. We set $|B_U| = |B_L|$. Learning rate for SGD is set according to this batch size. On CIFAR-10, for $|B_L| < 50$ we set the learning rate to $0.1$ and for $|B_L| = 50$ we set the learning rate to $0.3$. On CIFAR-100 and Mini-ImageNet we always have $|B_L| = 50$ and we set the learning rates to $0.5$ and $1.0$ respectively. We decay the learning rate using cosine annealing so that the learning rate would be 0 after 2000 rounds.

\myparagraph{\method parameters } We set the LSH dimension to $L=4096$ as this gave near exact approximation of the cosine similarities while still being computationally fast (less than 1 second per round). We set the sparsification parameter to $k=10$, so that each point is connected to its 10 most similar neighbors in the graph, and the label propagation parameter to $\alpha =0.99$.

\section{Additional Experiments}

\subsection{XCLP for Prediction}

We include extra experiments to test the performance of \method for prediction on an additional dataset. We follow the notation and setup detailed in Section \ref{sec:LPpred}.

\paragraph{Dataset} We use the FEMNIST \cite{leaf} dataset. FEMNIST is a federated dataset for handwritten character recognition, which has $62$ classes (digits and lower/upper case letters). It has 817,851 samples and we keep the natural partition into 3597 clients based on the writer that wrote each character. The clients are non-\iid, as they heterogeneous in the amount of data they possess, the classes they have, as well as the data distributions themselves. 

\paragraph{Experimental Setup}
We consider a setting where each client possesses partly labeled training data and wishes to infer labels for their own unlabeled new data. We looks at a range of different labeling scenarios based on what fraction, $\alpha$, of each client's training data is labeled. Furthermore, due to the large number of clients and datapoints in FEMNIST we reduce communication overhead by partitioning the clients into large groups (we use 50 groups, each with approximately 700 clients) and run \method separately on each groups of clients. Per-client LP remains as described in \ref{sec:LPpred}.

We investigate two different choices for the preprocessing function $h$. For the first we simply use the identity (\ie no embedding) and run both per-client LP and \method directly on the raw data. For the second we use a linear embedding that we trained over the client data. Specifically, following the low training overhead approach of \ref{sec:LPpred}, we train a two layer linear MLP using Federated Averaging and use the first layer linear embedding as our preprocessing function.

\paragraph{Results}

Table \ref{tab:femnist_results} reports the results obtained when different fractions $\alpha$ of each client's training data are labeled. The left columns ("identity") give results for the one-shot setting, where no training is required, and LP is run directly on the client features. The right columns ("linear") illustrate that by embedding the features using a simple linear layer we are able to improve the performance of \method. In all cases \method substantially outperforms per-client LP. The difference is even more noticeable than in \ref{sec:LPpred}, presumably due to the smaller amount of data present at each client.

\begin{table*}[t]

\caption{\method for prediction (FEMNIST dataset): Classification accuracy [in \%] with 
two different preprocessing functions (identity and linear embedding) and different fractions of labeled data available (average and standard deviation across three runs).
}
\label{tab:femnist_results}
\begin{center}
\small
\begin{tabular}{|c||c|c||c|c|c|}\hline
 & \multicolumn{2}{c||}{identity} & \multicolumn{3}{c|}{linear} \\
\textit{Fraction of labeled data} & \emph{per-client LP} & \emph{\method} & \emph{FedAvg} & \emph{per-client LP} & \emph{\method} \\
 \hline\hline
 $\alpha=0.1$ & $33.74 \pm 0.20$ & $\bf49.29 \pm 0.16$ & $\bf53.09 \pm 0.19$ & $37.27 \pm 0.12$ & $52.39 \pm 0.30$
 \\\hline
 $\alpha=0.2$ & $44.79 \pm 0.03$ & $\bf53.93 \pm 0.04$ & $57.54 \pm 0.46$ & $50.89 \pm 0.11$ & $\bf58.57 \pm 0.13$
 \\\hline
 $\alpha=0.5$ & $50.99 \pm 0.06$ & $\bf59.37 \pm 0.13$ & $61.07 \pm 0.49$ & $59.20 \pm 0.19$ & $\bf63.46 \pm 0.11$
 \\\hline
 $\alpha=1.0$ & $51.91 \pm 0.07$ & $\bf62.35 \pm 0.07$& $62.90 \pm 0.31$ & $59.42 \pm 0.11$ & $\bf66.18 \pm 0.20$
\\\hline
\end{tabular}
\end{center}
\end{table*}

\section{Additional Background}\label{app:background}

\subsection{Propagating values along a data graph} \label{app:LP}
For a graph with $n$ vertices and adjacency matrix 
$\mathcal{W}\in\R^{n\times n}$ of edge weights, a vector of values, 
$y\in\R^n$, can be propagated to neighboring vertices by forming 
$z=\mathcal{W}y$. 
By repeatedly multiplying with $\mathcal{W}$ values can be propagated 
all along the graph structure~\citep{zhu2005semi}. 

In the context of propagating labels, one wants not only to propagate 
the labels to unlabeled points, but also to prevent the information at 
labeled points to be forgotten. For that, one can uses an extended update 
rule
\begin{equation}
    z_{t+1} = \alpha \mathcal{W} z_t + y   \label{eq:z_update}
\end{equation}
where $\alpha \in (0, 1)$ is a trade-off hyperparameter. 
For $\|\mathcal{W}\|<1/\alpha$, this process has the closed form expression 
\begin{equation}
    z_{\infty} = (\operatorname{Id} - \alpha \mathcal{W})^{-1} y,  \label{eq:z_limit}
\end{equation}
as its ($t\to\infty$)-limit. This can be seen from the fact that \Eqref{eq:z_update} 
is a contraction with $z_{\infty}$ as fixed point.

In Section~\ref{subsec:method} we make use of this fact together with the 
observation that \Eqref{eq:z_limit} can readily be applied to vectors-valued 
data with a matrix $Y$ in place of $y$. 
However, the propagation step will not preserve normalization, 
\eg of the $L^1$-norm.
When such a property is required, \eg for the calculation of entropy-based confidence,
normalization has to be performed explicitly post-hoc.

\subsection{Computing similarity from hashed data} \label{app:LSH}
Locality-sensitive hashing (LSH)~\citep{indyk1998approximate} is a procedure 
for hashing real-valued vectors into binary vectors while preserving their 
pairwise similarity. 
Let $v\in\R^d$ be a vector. To encode $v$ into a binary vector $b$ of length $L$, 
LSH randomly samples $L$ hyperplanes in $\R^d$. For each hyperplane it checks 
whether $v$ lies above or below it and sets the $i$th bit in $b$ as $1$ or $0$ accordingly.
Formally, $b_i = \mathbbm{1}_{\langle v, u_i \rangle\geq 0}$, where 
$u_i \in \R^d$ is the normal vector of the $i$th hyperplane. 
A key property of LSH is that it approximately preserves cosine-similarity.
Concretely, for vectors $v_1,v_2$ with LSH encodings $b_1, b_2$ (compute with
the same projections), one has 
\begin{equation}
    \frac{\langle v_1, v_2\rangle}{\|v_1\|\|v_2\|} \approx \cos( \pi h(b_1, b_2)/L)
    \label{eq:hamming_cosine}
\end{equation}
where $h$ is the Hamming distance (number of bits that differ) between two binary 
vectors. 
The reason is that the probability of $b_1$ and $b_2$ differing at 
any bit $i$ is the probability that the $i$-th sampled hyperplane 
lies between $v_1$ and $v_2$, which is equal to $\measuredangle(v_1,v_2)/\pi$. 
%see \Cref{subfig:LSH}. % for an illustration.
%
By the law of large numbers, the more hyperplanes one samples, the better the approximation quality.

\end{document}